\documentclass[10pt,twocolumn,letterpaper]{article}
\usepackage[
  backend=bibtex,
   style=ieee,
  citestyle=numeric-comp,
  natbib=true,
  mincitenames=1,
  maxcitenames=1,
]{biblatex}
\bibliography{egbib}
\usepackage{cvpr}
\usepackage{times}
\usepackage{epsfig}
\usepackage{graphicx}
\usepackage{amsmath}
\usepackage{amssymb}

\AddToShipoutPicture{%
  \AtTextUpperLeft{%
    \put(0,\LenToUnit{1cm}){%
      \fbox{\parbox{\textwidth}{%
        \centering
        This work was accepted to be presented at the IEEE/ISPRS Workshop on Large Scale Computer Vision for Remote Sensing Imagery (EarthVision) to be held at the IEEE Conference on Computer Vision and Pattern Recognition (CVPR) 2019.
      }}%
    }%
  }%
}%

\usepackage{subcaption}
\usepackage{siunitx}
\DeclareSIUnit\px{px}

\usepackage[%
  breaklinks=true,
  letterpaper=true,
  colorlinks,
  bookmarks=false,
]{hyperref}

\usepackage[%
  xindy,
  acronym,
]{glossaries}
\glsdisablehyper
\defglsdisplayfirst[acronym]{\emph{#1#4}}

\newacronym{gl:DSM}{DSM}{digital surface model}
\newacronym{gl:nDSM}{nDSM}{normalized digital surface model}
\newacronym{gl:DTM}{DTM}{digital terrain model}
\newacronym{gl:GPU}{GPU}{graphics processing unit}
\newacronym{gl:GSD}{GSD}{ground sampling distances}
\newacronym{gl:cLSGAN}{cLSGAN}{conditional least square generative adversarial network}
\newacronym{gl:cGAN}{cGAN}{conditional generative adversarial network}
\newacronym{gl:GAN}{GAN}{generative adversarial network}\newacronym{gl:LoD}{LoD}{level of detail}
\newacronym{gl:CityGML}{CityGML}{city geography markup language}
\newacronym{gl:CNN}{CNN}{convolutional neural network}
\newacronym{gl:SGM}{SGM}{semi-global matching}
\newacronym{gl:SGD}{SGD}{stochastic gradient descent}
\newacronym{gl:BN}{BN}{batch normalization}
\newacronym{gl:ReLU}{ReLU}{rectified linear unit}
\newacronym{gl:LReLU}{LReLU}{leaky rectified linear unit}
\newacronym{gl:LIDAR}{LIDAR}{light detection and ranging}
\newacronym{gl:PAN}{PAN}{pan-chromatic}

\newcommand{\V}[1]{\ensuremath{\boldsymbol{#1}}}
\newcommand{\HybridcGANdot}{Hybrid-\glsname{gl:cGAN}}
\newcommand{\HybridcGAN}{\HybridcGANdot\space}
\newcommand{\WNetdot}{WNet}
\newcommand{\WNet}{\WNetdot\space}
\newcommand{\WNetcGANdot}{WNet-\glsname{gl:cGAN}}
\newcommand{\WNetcGAN}{\WNetcGANdot\space}

\newcommand{\e}{\ensuremath{\mathrm{e}}}
\newacronym{gl:RMSE}{RMSE}{root mean squared error}
\newacronym{gl:MAE}{MAE}{mean absolute error}
\newacronym{gl:NMAD}{NMAD}{normalized median absolute deviation}
\newacronym{gl:SVM}{SVM}{support vector machine}
\newacronym{gl:IDW}{IDW}{inverse distance weighting}
\newacronym{gl:CRF}{CRF}{conditional random field}

\renewcommand{\V}[1]{\ensuremath{\boldsymbol{#1}}}
\newcommand{\generator}{\ensuremath{G}\xspace}
\newcommand{\discriminator}{\ensuremath{D}\xspace}
\newcommand{\latentZ}{\ensuremath{z}\xspace}
\newcommand{\latentZVec}{\ensuremath{\V{\latentZ}}\xspace}
\newcommand{\targetY}{\ensuremath{y}\xspace}
\newcommand{\targetYVec}{\ensuremath{\V{\targetY}}\xspace}
\newcommand{\inputX}{\ensuremath{x}\xspace}
\newcommand{\inputXVec}{\ensuremath{\V{\inputX}}\xspace}

\newcommand{\errorRMSE}{\ensuremath{\varepsilon_\text{\acrshort{gl:RMSE}}}\xspace}
\newcommand{\errorNMAD}{\ensuremath{\varepsilon_\text{\acrshort{gl:NMAD}}}\xspace}
\newcommand{\errorMAE}{\ensuremath{\varepsilon_\text{\acrshort{gl:MAE}}}\xspace}

\usepackage[capitalize]{cleveref} 
\cvprfinalcopy % *** Uncomment this line for the final submission

\cvprfinalcopy
\usepackage{booktabs}
\usepackage{array}
\newcolumntype{x}{l}
\newcolumntype{X}{>{\scriptsize}l}
\newcolumntype{v}[1]{>{\raggedright\hspace{0pt}}p{#1}}
\newcolumntype{V}[1]{>{\scriptsize\raggedright\hspace{0pt}}p{#1}}

\usepackage{xcolor}

\makeatletter
\@namedef{ver@everyshi.sty}{}
\makeatother
\usepackage{tikz}

\def\cvprPaperID{38} % *** Enter the CVPR Paper ID here

\ifcvprfinal\fi
\begin{document}

%%%%%%%%% TITLE
\title{Late or Earlier Information Fusion from Depth and Spectral Data? Large-Scale Digital Surface Model Refinement by \HybridcGAN}
%\titleheader{Manuscript is submitted to Computer Vision and Pattern Recognition Workshop}

\author{Ksenia Bittner, Peter Reinartz\\
German Aerospace Center (DLR)\\
Munich, Germany\\
{\tt\small ksenia.bittner@dlr.de}\\
{\tt\small peter.reinartz@dlr.de}
% For a paper whose authors are all at the same institution,
% omit the following lines up until the closing ``}''.
% Additional authors and addresses can be added with ``\and'',
% just like the second author.
% To save space, use either the email address or home page, not both
\and
Marco K\"orner\\
Technical University of Munich (TUM)\\
Munich, Germany\\
{\tt\small marco.koerner@tum.de}
%\and
%Peter Reinartz\\
%German Aerospace Center (DLR)\\
%Munich, Germany\\
%{\tt\small peter.reinartz@dlr.de}
}

\maketitle
\thispagestyle{empty}

%%%%%%%%% ABSTRACT
\begin{abstract}
We present the workflow of a \gls{gl:DSM} refinement methodology using a \textsc{\HybridcGAN} where the generative part consists of two encoders and a common decoder which blends the spectral and height information within one network.
The inputs to the \HybridcGAN are single-channel photogrammetric \glspl{gl:DSM} with continuous values and single-channel \gls{gl:PAN} half-meter resolution satellite images. 
Experimental results demonstrate that the earlier information fusion from data with different physical meanings helps to propagate fine details and complete an inaccurate or missing 3D information about building forms.
Moreover, it improves the building boundaries making them more rectilinear.
\end{abstract}

%%%%%%%%% BODY TEXT
\glsresetall
\section{Introduction}

\Glspl{gl:DSM} with detailed and realistic building shapes are beneficial data sources for many remote sensing applications, such as urban planning, cartographic analysis, environmental investigations, \etc.
Several ways for deriving 3D elevation morphology, but the \glspl{gl:DSM} generated from space-borne data using stereo image matching techniques nowadays have lower costs compared to other technologies, show relatively high spatial resolution and wide coverage which is important for large-scale remote sensing applications, such as disaster monitoring. 

\begin{figure}[t]
  \centering
  \subcaptionbox{\label{fig:PAN2D}}
                {\frame{\includegraphics[width=.3\linewidth]{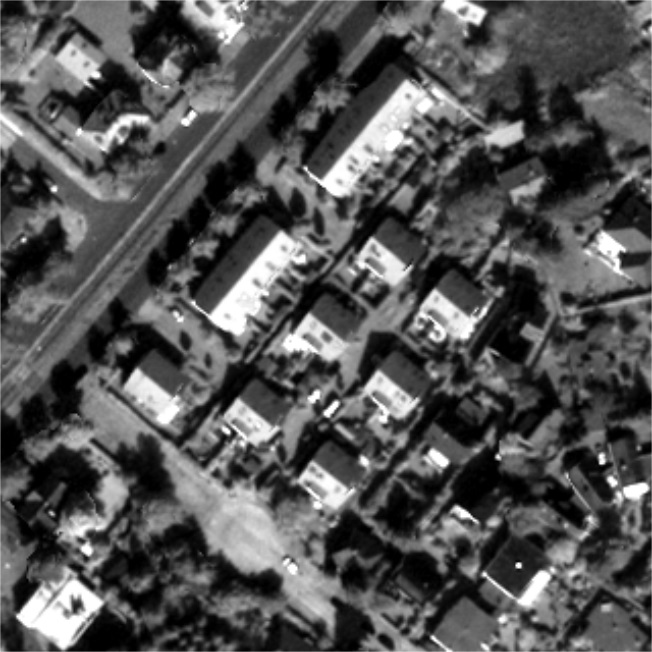}}
                }
  \subcaptionbox{\label{fig:Dfilled2D}}
                {\frame{\includegraphics[width=.3\linewidth]{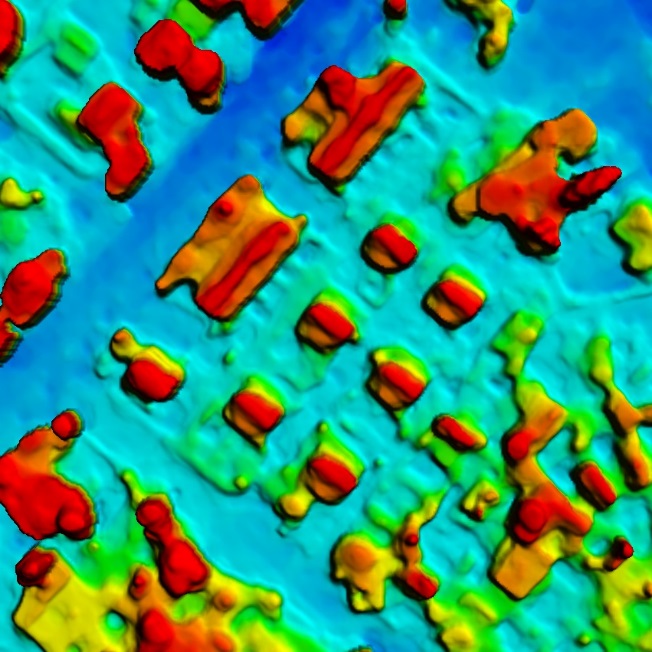}}
                }  
  \subcaptionbox{\label{fig:GT2D}}
                {\frame{\includegraphics[width=.3\linewidth]{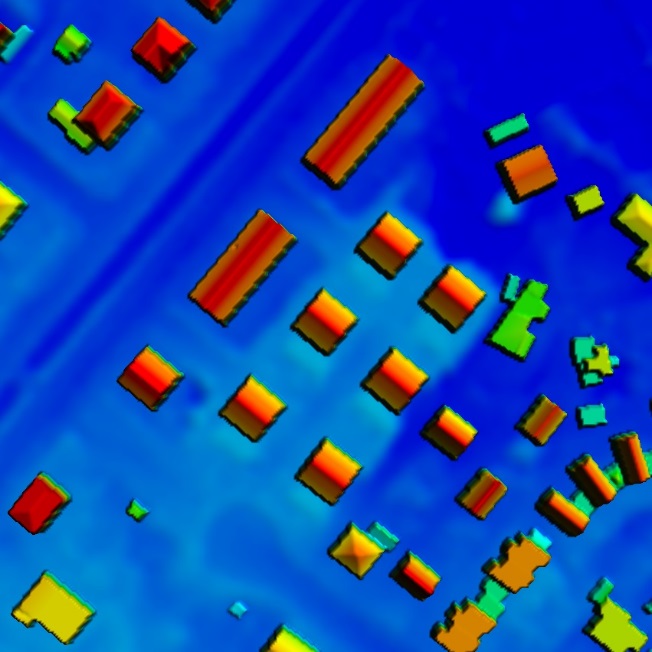}}
                }
  \caption{ Sample of area from our dataset illustrated both inputs to the network   
  \subref{fig:PAN2D} \gls{gl:PAN} image and \subref{fig:Dfilled2D}
  photogrammetric \gls{gl:DSM}, and \subref{fig:GT2D} the ground truth \gls{gl:LoD}-2-\gls{gl:DSM}.
  The \gls{gl:DSM} images are color-shaded for better visualization. \label{fig:Img1}}
\end{figure}

However, apart from many advantages, \glspl{gl:DSM} generated by image-based matching techniques show a reasonable amount of noise and outliers because of matching errors or occlusions due to densely located buildings or trees, which cover parts of the building constructions. 
To solve these problems, algorithms from computer vision have been analyzed and adapted to satellite imagery processing.
In the literature, very few of the proposed approaches work towards photogrammetric \gls{gl:DSM} improvement of urban areas. 
In contrast, earlier methodologies investigate the \glspl{gl:DSM} refinement by
applying filtering techniques, such as Gaussian~\cite{walker2006comparative}, 
Kalman~\cite{wang1998applying}, geostatistical~\cite{felicisimo1994parametric}, or interpolation routines, including \gls{gl:IDW} and kriging~\cite{anderson2005lidar}.
Despite achieving smoother surfaces of roof planes, they negatively influence the steepness of the building walls. 
Later, some methodologies proposed to additionally utilize spectral images for \gls{gl:DSM} refinement tasks, as, for instance, they contain accurate information about object boundaries or texture. 
For instance, \citet{krauss2010enhancement} transfer segmentation information from stereo satellite imagery to the \gls{gl:DSM} and, from statistical analysis and spectral information, perform object detection and classification. 
As a further step, this information is used to refine the \gls{gl:DSM}.
\Citet{sirmacek2010enhancing} propose building shape refinement on \glspl{gl:DSM} by a multiple-step procedure. 
First, they extract possible building segments by thresholding the \gls{gl:nDSM}. 
Then, by considering Canny edge information from the spectral image, they fit rectangular boxes to detect building shapes.
Finally, applying this information about detected rectangular boundaries, the building shapes are further enhanced.
The drawback of this methodology is that they assign one single height value to each generated building object.
Moreover, it is limited to the detection and enhancement of rectangular buildings only.

\vspace{-5pt}

Within recent years, the already well-studied branch of the deep learning family, \glspl{gl:CNN}, has also been applied to the remote sensing field. 
They achieve state-of-the-art results for image classification, object detection, or semantic segmentation tasks. 
However, most of these methodologies extract information from spectral imagery, while depth image processing is still not well explored, as it is not straightforward to work with continuous values.
For example,~\citet{eigen2014depth} employ cascade deep network which first performs a coarse global prediction from a single spectral image and refines the predictions locally afterwards.
\Citet{liu2016learning} join \glspl{gl:CNN} and \glspl{gl:CRF} in a unified framework while making use of superpixels for preserving sharp edges.
In case of height prediction problems, only a couple of works made attempts to improve \glspl{gl:DSM}.
%~\Citet{li2017single} investigate a novel set loss and a two-streamed \gls{gl:CNN} that merges predictions of depth and depth gradients.
%, Xu et al. [32] propose to integrate complementary informationderived from multiple CNN side outputs using CRFs.
Our earlier approaches~\cite{Bittner18:ALSb,bittner2018dsm} propose to refine building shapes to a high level of details from photogrammetric half-meter resolution satellite \glspl{gl:DSM} using \gls{gl:GAN}-based techniques. 
Mainly, they investigate \glspl{gl:cGAN} with two objective functions---\ie the negative log-likelihood and least square residuals---to generate accurate \gls{gl:LIDAR}-like and \gls{gl:LoD}-2-like \glspl{gl:DSM} with enhanced 3D building shapes directly from noisy input data.  
Moreover, the experiment performed on a completely new dataset, belonging to different geographical areas, showed the network's potential to generalize well to different cities with complex constructions without many difficulties.
As it is common in the field of remote sensing to fuse data with different modalities to complement missing knowledge, in our followed work \cite{Bittner19:new} we introduce a \gls{gl:cGAN}-based network which merges depth and spectral information within an end-to-end framework.
Fusing data from separate networks---which are fed with \gls{gl:PAN} images and \glspl{gl:DSM}---is performed at a \emph{later} stage right before producing the final output.

Following up to the aforementioned approach,
% \citet{Bittner19:new}, 
in this work, we investigate the fusion of spectral (\cref{fig:PAN2D}) and height (\cref{fig:Dfilled2D}) information at an \emph{earlier} stage within an end-to-end \HybridcGAN network to further improve not only the small and simple residential buildings, but also complex industrial ones. 
Besides, we add an auxiliary \emph{normal vector loss} term to the objective function enforcing the model to produce more planar and flat roof surfaces, similar to the desired \gls{gl:LoD}2-\gls{gl:DSM} (\cref{fig:GT2D}) artificially produced from \gls{gl:CityGML} data.

%As \glspl{gl:GAN} have been proven to be suitable for 3D data generation, in this work we  to refine building shapes on \glspl{gl:DSM} by combining spectral and height information within an end-to-end \gls{gl:GAN} network.

\section{Methodology}

\subsection{Objective function}

The advent of \gls{gl:GAN}-based domain adaptation neural networks,
introduced by \citet{goodfellow2014generative},
lead to great performance gains in generating realistic, but entirely artificial images.
These \glspl{gl:GAN} realize an adversarial manner of learning by training a pair of networks in a competing way: 
A \emph{generator} $\generator(\latentZVec| \inputXVec) = \targetYVec$ 
% tries to trick the discriminator by generating 
produces a fake image $\targetYVec$ from a latent vector $\latentZVec \sim {p}_\latentZ(\cdot)$ drawn from any distribution ${p}_\latentZ(\cdot)$%
---\eg a uniform distribution ${p}_\latentZ(\cdot) = \operatorname{Unif}(\cdot; a,b)$---%
which looks like a real one $\targetYVec^* \sim {p}_\text{real}(\cdot)$.
Adversarially, a \emph{discriminator} $\discriminator(\targetYVec| \inputXVec) \in [0,1]$ tries to decide whether a presented image $\targetYVec$ is a real or a generated fake one.
Frequently, some external information in the form of a source image \inputXVec is additionally used as an input to restrict both the generator in its output and the discriminator in its expected input. 
The objective function for such conditional \glspl{gl:GAN} can be expressed through a two-player minimax game
\begin{align}
  \underset{\generator}{\min} \; \underset{\discriminator}{\max} \; &\mathcal{L}_\text{cGAN}(G,D) = 
  \mathbb{E}_{\inputXVec,\targetYVec \sim p_\text{real}(\targetYVec)}[\log \discriminator(\targetYVec| \inputXVec)] + \nonumber \\ 
  &\mathbb{E}_{\inputXVec, \latentZVec \sim p_\latentZ(\latentZVec)}[\log(1 - \discriminator(\generator(\latentZVec | \inputXVec)| \inputXVec))] ,
  \label{eq:cGAN_objective}
\end{align}
where $\mathbb{E}[\cdot]$ denotes the expectation value.
To overcome the problem of instability during training \gls{gl:cGAN} when its objective function is based on the negative log-likelihood, we use a least square loss $L_2$ instead which yields the \gls{gl:cLSGAN} objective function
\begin{align}
  \mathcal{L}_\text{cLSGAN}(\generator,\discriminator) 
  &= \mathbb{E}_{\inputXVec,\targetYVec \sim p_\text{real}(\targetYVec)}[(\discriminator(\targetYVec| \inputXVec) - 1)^2] \nonumber \\ 
  &+ \mathbb{E}_{\inputXVec, \latentZVec \sim p_\latentZ(\latentZVec)}[\discriminator(\generator(\latentZVec | \inputXVec)| \inputXVec)^2].
  \label{eq:cLSGAN_objective}
\end{align}
In order to obtain \glspl{gl:DSM} in which the buildings feature sharply defined ridgelines and steep walls, we utilize the $L_1$ distance
\begin{align}
  \mathcal{L}_{L_1}(\generator) &= \mathbb{E}_{\inputXVec,\targetYVec \sim p_\text{real}(\targetYVec), \latentZVec \sim p_\latentZ(\latentZVec)}[\lVert \V{y} - G(\latentZVec | \inputXVec) \rVert_1],
  \label{eq:L1}
\end{align}
which prevents blurring effects.

As our major goal is to improve roof surfaces by making them flatter and looking closer to realistic ones, we integrate a \emph{normal vector loss} term
\begin{align}
\label{eq:surface_normal}
\mathcal{L}_{\text{normal}}(\mathcal{N}^\text{t}, \mathcal{N}^\text{p}) 
&= \frac{1}{m} \sum_{i=1}^{m} 
   \left( 1 - \frac{\left\langle \V{n}_{i}^\text{t}, \V{n}_{i}^\text{p} \right\rangle } 
                   {\left\| \V{n}_{i}^\text{t} \right\| \left\| \V{n}_{i}^\text{p}\right\| } \right),
\end{align}
into the learning process, as proposed by \citet{hu2018revisiting}.
This normal vector loss measures the angle between the set of surface normals 
$\mathcal{N}^\text{p} = \left\{ \V{n}_1^\text{p}, \ldots, \V{n}_m^\text{p}\right\}$ 
of an estimated \gls{gl:DSM} 
and the set of surface normals
$\mathcal{N}^\text{t} = \left\{ \V{n}_1^\text{t}, \ldots, \V{n}_m^\text{t}\right\}$
of the target \gls{gl:LoD}2-\gls{gl:DSM}.
Adding those three terms together leads to our final objective
\begin{align}
  \label{eq:objective_function}
  \generator^\star = \arg \underset{G}{\min} \; \underset{D}{\max} \; \mathcal{L}_{\text{cGAN}}(G,D)+\lambda \mathcal{L}_{L_1}(G) \nonumber \\
 + \gamma \mathcal{L}_{\text{normal}}(\mathcal{N}^\text{t}, \mathcal{N}^\text{p}) \,,
\end{align}
where \generator intents to minimize the objective function $\mathcal{L}_{\text{cLSGAN}}(G,D)$ against \discriminator that aims to maximize it.
The parameters $0 \leq \lambda \in \mathbb{R}$ and $0 \leq \gamma \in \mathbb{R}$ are the balancing hyper-parameters.

\begin{figure}[t]
 \centering
  %\hspace{-0.1cm}
  \includegraphics[width=\columnwidth]{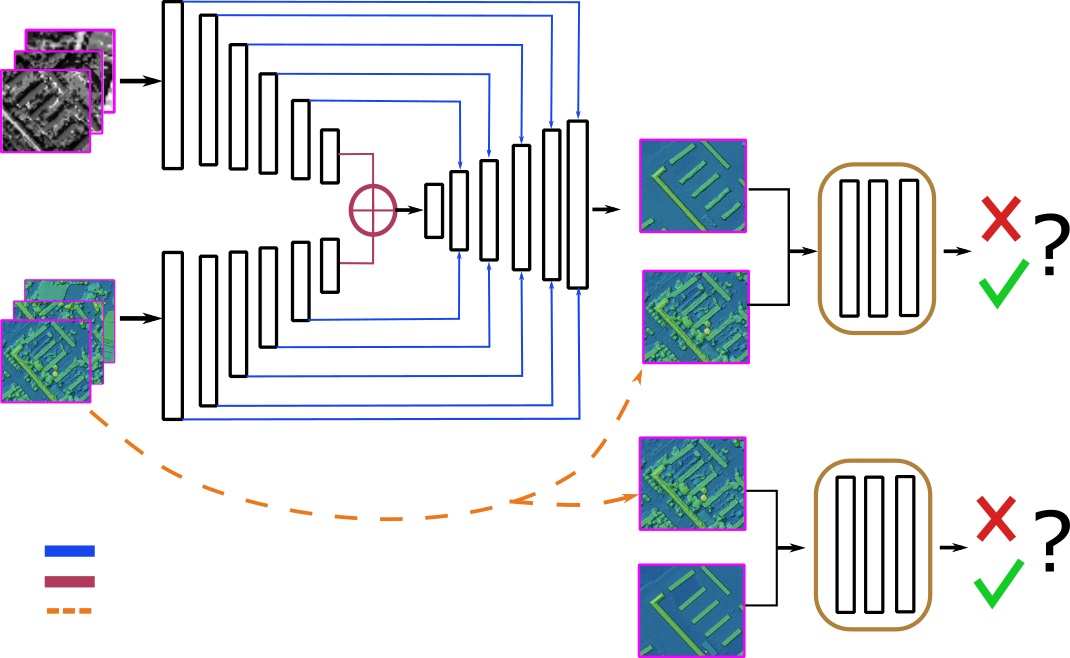}
  \put (-240,49) {\scriptsize{DSM ($I_1$)}}
  \put (-240,105) {\scriptsize{PAN} ($I_2$)}
  \put (-47,114) {\scriptsize{\discriminator}}
  \put (-47,48) {\scriptsize{\discriminator}}
  \put (-155,43) {\scriptsize{\generator}}
  \put (-100,118) {\scriptsize{\generator($I_1,I_2$)}}
  \put (-90,58) {\scriptsize{$I_1$}}
  \put (-200,150) {\scriptsize{$E_2$}}
  \put (-200,93) {\scriptsize{$E_1$}}
  %\put (-165,23) {\scriptsize{conditioning}}
  %\put (-90,118) {\scriptsize{$I_1$}}
  \put (-90,50) {\scriptsize{$I_1$}}
  \put (-90,-6) {\scriptsize{GT}}
  \put (-212,23) {\scriptsize{skip connections}}
  \put (-212,16) {\scriptsize{concatenation}}
  \put (-212,10) {\scriptsize{condition}}
  %\put (-39,116) {\textbf{\scriptsize{\discriminator}}}
  %\put (-207,58) {\rotatebox{90}{\color{black}\textbf{\scriptsize{Fusion}}}}
  %\put (-187,70) {\rotatebox{90}{\color{black}\textbf{\scalebox{.5}{conv $1\times 1$}}}}
  \caption{Schematic overview of the proposed architecture for the building shape refinement on photogrammetric \glspl{gl:DSM} by \HybridcGAN using both depth and spectral information.}
  \label{fig:architecture} 
\end{figure}
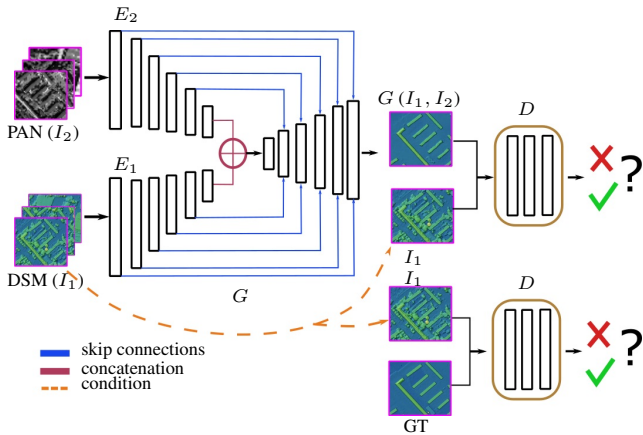

\subsection{Network architecture}

We have already made an attempt to investigate the \gls{gl:LoD}2-like \gls{gl:DSM} generation with enhanced building shapes from a single noisy photogrammetric \gls{gl:DSM} using \gls{gl:cGAN} architecture with an objective function based on least square residuals~\cite{bittner2018dsm}.
In this work, we refer to it as a single-stream \gls{gl:cGAN}. 
%Recently, it has been already experimented with good-quality elevation model generation from inaccurate and noisy photogrammetric \glspl{gl:DSM} on the basis of \gls{gl:cGAN} architectures proposed by~\citet{isola2016image}. 
As each photogrammetric \gls{gl:DSM} is a product obtained from panchromatic image pairs, it seems reasonable to integrate depth and spectral data into one single network, as the latter provides sharper information about building silhouettes, which allows not only a better reconstruction of missing building parts but also the refinement of building outlines. 
We fuse two separate but identical \emph{UNet}-type networks at the later end within the \generator part of a \gls{gl:cGAN}, where the first stream is fed with the \gls{gl:PAN} image and the second stream with the stereo \gls{gl:DSM}, creating a so-called \WNet architecture~\cite{Bittner19:new}.
%The \discriminator construction is kept the same as in previous works~\cite{bittner2018automatic,bittner2018dsm}. 
% The detailed description of the \WNetcGAN architecture can be found in the original paper published by \citet{Bittner19:new}.

In this paper, we examine the potential of an earlier fusion of data from different modalities, as it could potentially even better blend together the depth and spectral information. 
Mainly, the generator \generator of the proposed \HybridcGAN network consists of two encoders $E_1$ and $E_2$, concatenated at the top layer, and a common decoder, which integrates information from two different modalities and generates an \gls{gl:LoD}2-like \gls{gl:DSM} with refined building shapes. 
The inputs to $E_1$ are the single-channel orthorectified \gls{gl:PAN} images, while $E_2$ receives the single-channel photogrammetric \glspl{gl:DSM} with continuous values.
Since intensity and depth information have different physical meanings, it is unlikely to make sense to jointly propagate them right from the beginning. 
It seems reasonable to separate them first and allow the network to learn the most valuable features from each modality itself.
%The generator \generator is constructed via U-form network with skip connections from both spectral-stream and depth-stream allowing the decoder to learn back detailed features that are lost due to the pooling in the encoders. 
%The generator \generator is constructed by a U-form network with 14 skip connections from both the spectral stream and the depth stream allowing the decoder to learn back detailed features that were lost by pooling in the encoders. 
The generator \generator is constructed based on an U-form network proposed by \citet{isola2016image}.
As a result, in our case it has 14 skip connections from both the spectral stream and the depth stream allowing the decoder to learn back detailed features that were lost by pooling in the encoders.
In particular, the encoder and decoder consist of 8 convolutional layers each, followed by a \gls{gl:LReLU}~\cite{Maas13:RNL} activation function
\begin{align*}
\sigma_\text{\acrshort{gl:LReLU}}(z)=
	\begin{cases}
	  z, & \text{if $z>0$}\\
    az, & \text{otherwise}
   \end{cases}, a \in \mathbb{R}^+,
\end{align*}
and \gls{gl:BN} in case of the encoder, and a \gls{gl:ReLU} activation function
\begin{align*}
\sigma_\text{\acrshort{gl:ReLU}}(z)=
	\begin{cases}
	  z, & \text{if $z>0$}\\
    0, & \text{otherwise}
   \end{cases}       
\end{align*}
and \gls{gl:BN} in case of the decoder.
On top of the generator network \generator, the $\operatorname{tanh}$ activation function $\sigma_\text{tanh}(z) = \operatorname{tanh}(z)$ is applied.

The discriminator network \discriminator is a binary classification network constructed with 5 convolutional layers, followed by \gls{gl:LReLU} activation function and a \gls{gl:BN} layer. 
It has a \emph{sigmoid} activation function $\sigma_\text{sigm}(z) = \frac{1}{1+\e^{-z}}$ at the top layer to output the likelihood that the input image belongs either to class 1 (\enquote{real}) or class 0 (\enquote{generated}).
A schematic representation of the proposed architecture is depicted in \cref{fig:architecture}.

All along the training phase, the two networks \generator and \discriminator are trained at the same time by alternating one gradient descent step of \discriminator and one gradient descent step of \generator. 
During the inference process, only the trained generator model \generator of the \HybridcGAN network is involved.

\section{Study Area and Experiments}

\subsection{Dataset}

Experiments have been carried out on data showing the city of Berlin, Germany, covering a total area of \SI{410}{\square\kilo\meter}.
It consists of half-meter resolution orthorectified \gls{gl:PAN} images showing the closest nadir view and photogrammetric \glspl{gl:DSM} generated with \gls{gl:SGM}~\cite{d2011semiglobal} from six panchromatic Worldview-1 images acquired at two different days.
%derived from six WorldView-1 half-meter resolution stereo panchromatic imagery.
%To fit remote sensing images into the available GPU memory, we tile \gls{gl:DSM} and PAN images into patches of size \SI{256x256}{\px} with corresponding area.
As ground truth, the artificial \gls{gl:LoD}2-\gls{gl:DSM}, generated with a resolution of \SI{0.5}{\px} from a \gls{gl:CityGML} data model together with an available \gls{gl:DTM}, was used.
The \gls{gl:CityGML} data model is freely available from the download portal \emph{Berlin 3D} \footnote{\url{http://www.businesslocationcenter.de/downloadportal}}.
We followed the same \gls{gl:LoD}2-\gls{gl:DSM} creation procedure as in \citet{bittner2018dsm}. 
% The detailed methodology on \gls{gl:LoD}2-\gls{gl:DSM} creation procedure is given by~\citet{bittner2018dsm}. 

\begin{figure*}[t]
  \centering
  \subcaptionbox{\label{fig:testArea_O3_width_320_height_320_2}\Acrlong{gl:PAN} image}
                {\includegraphics[width=.47\linewidth]{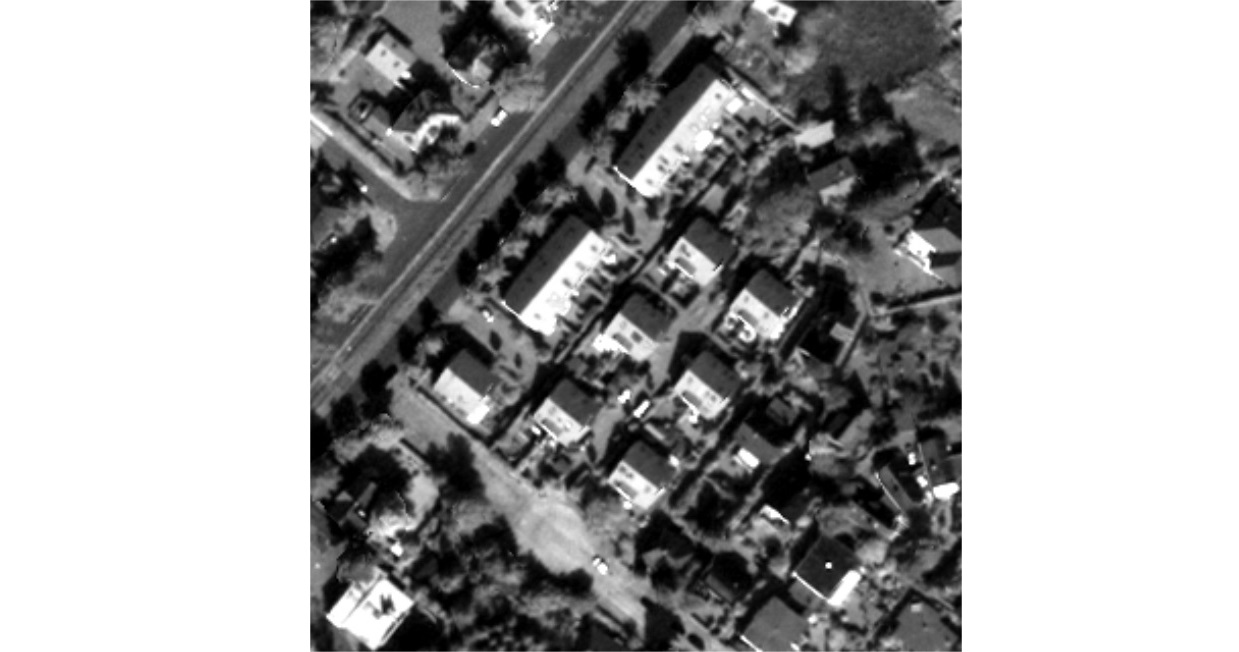}
                }\hfill
  \subcaptionbox{\label{fig:test_DSM2LOD2GEOKOLref_width_320_height_320} Single depth-stream \gls{gl:DSM}~\cite{bittner2018dsm}}
                {\includegraphics[width=.47\linewidth]{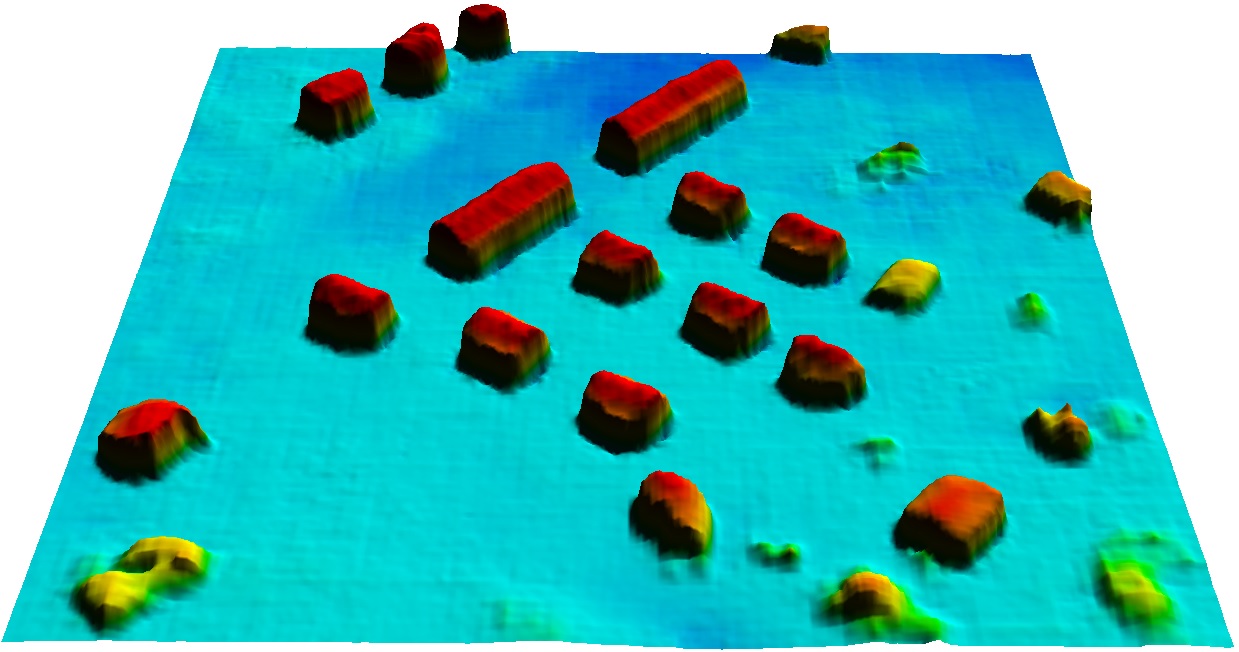}
                \linethickness{0.5mm}
                \put(-43,95){\color{magenta}\vector(-1, 0){14}}
                \put(-3,40){\color{magenta}\vector(-1, 0){14}}
                \put(-40,45){\color{magenta}\vector(-1, 0){14}}
                \put(0,20){\color{magenta}\vector(-1, 0){14}}
                }

  \subcaptionbox{\label{fig:testArea_Dfilled_geoid_width_320_height_320} Photogrammetric \gls{gl:DSM}}
                {\includegraphics[width=.47\linewidth]{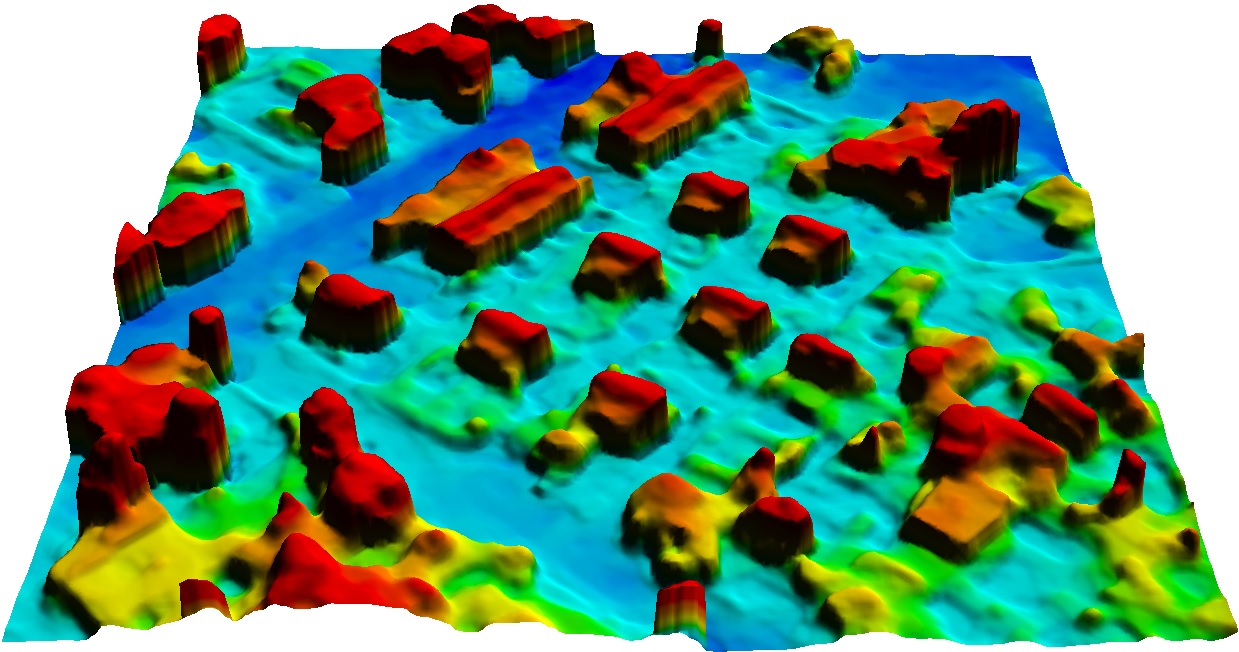}
                }\hfill
  \subcaptionbox{\label{fig:test_O3andDSM2LODGEOKOLref_lsgan_width_320_height_320} \WNetcGAN\cite{Bittner19:new}}
                {\includegraphics[width=.47\linewidth]{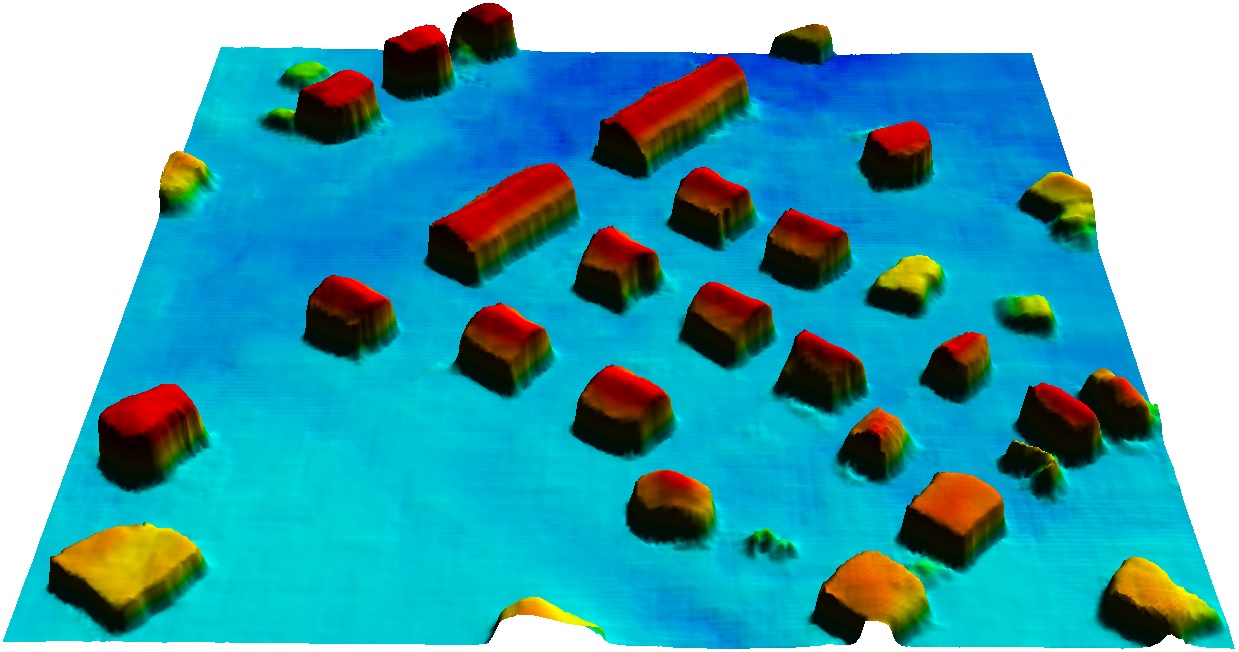}
                }

  \subcaptionbox{\label{fig:testArea_LOD2filled_width_320_height_320} Ground truth}
                {\includegraphics[width=.47\linewidth]{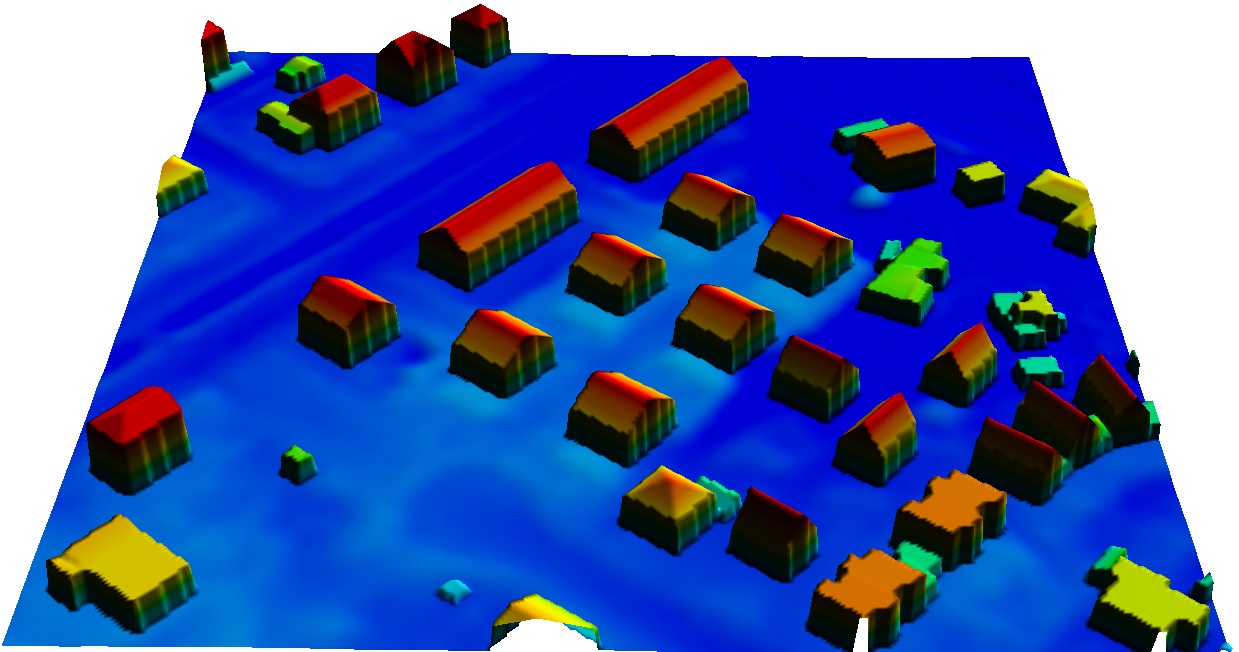}
                }\hfill
  \subcaptionbox{\label{fig:test_O3andDSM2LODGEOKOLref_lsgan_NotSharedWeights_width_320_height_320} \HybridcGAN (ours)}
                {\includegraphics[width=.47\linewidth]{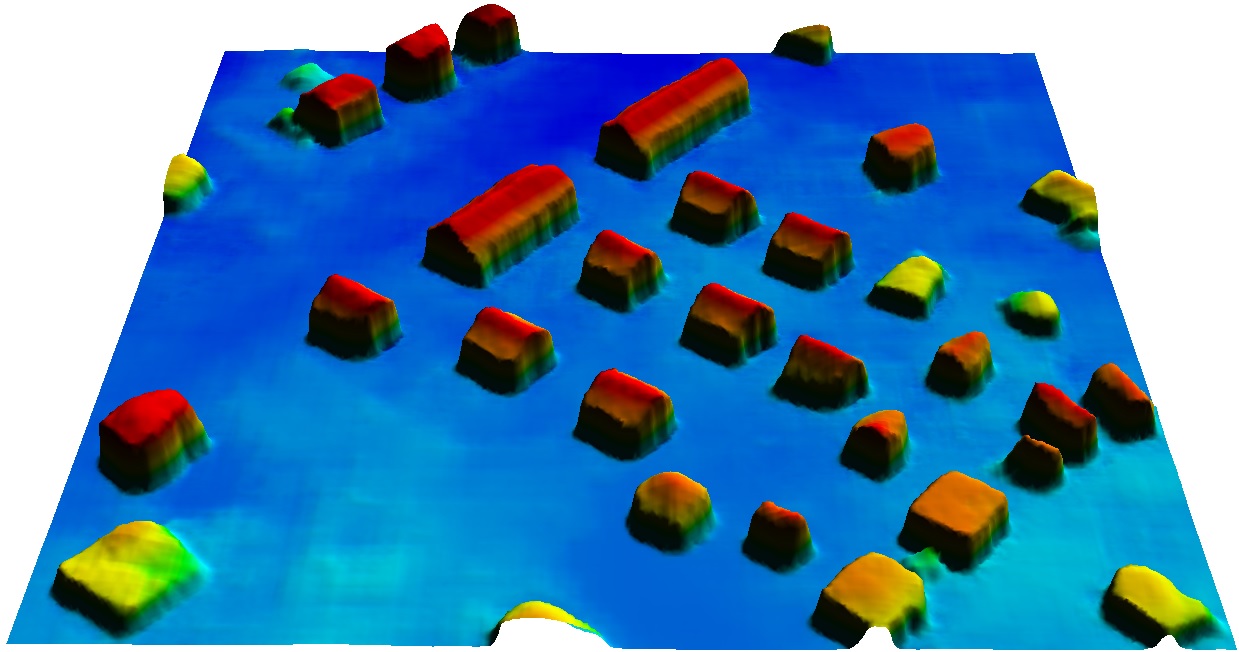}
                \put (-153,83) {\color{yellow}\textbf{\scriptsize{1}}}
                \put (-120,81) {\color{yellow}\textbf{\scriptsize{2}}}
                }
  \caption{%
    Visual analysis of \glspl{gl:DSM}, generated by
    \subref{fig:testArea_Dfilled_geoid_width_320_height_320} a standard photogrammetric method,
    \subref{fig:test_DSM2LOD2GEOKOLref_width_320_height_320} the single-stream \gls{gl:cGAN} model~\cite{bittner2018dsm}, 
    \subref{fig:test_O3andDSM2LODGEOKOLref_lsgan_width_320_height_320} the two-stream \WNetcGAN\cite{Bittner19:new},
    and \subref{fig:test_O3andDSM2LODGEOKOLref_lsgan_NotSharedWeights_width_320_height_320} our proposed \HybridcGAN architecture. 
    The \gls{gl:DSM} images are color-coded for better visualization.
  }
  \label{fig:smallBuildings}
\end{figure*}

\begin{figure*}[t]
  \centering
  
  \subcaptionbox{Single depth-stream \gls{gl:DSM}~\cite{bittner2018dsm}\label{fig:cLSGAN_profile}}
                {\parbox{.3\linewidth}{%
                   \includegraphics[width=\linewidth]{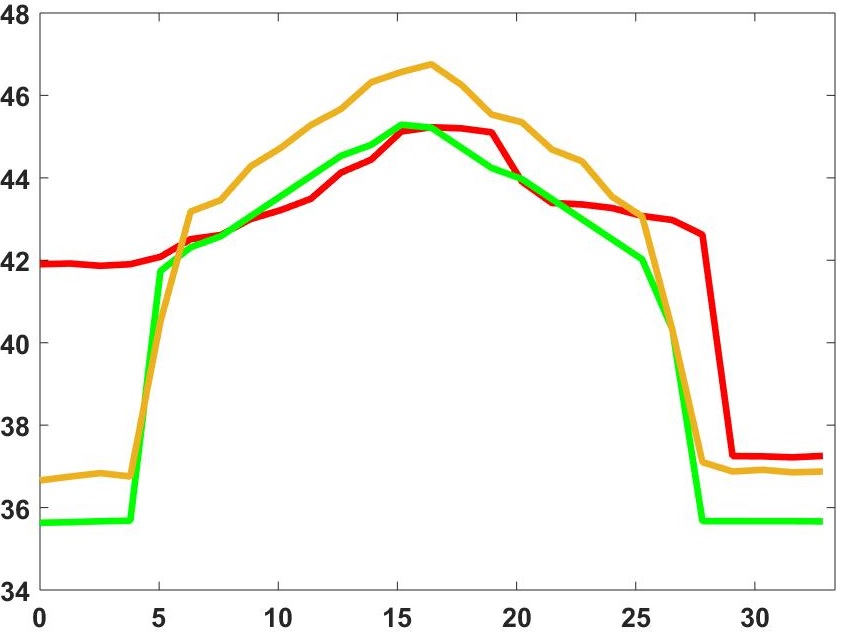}\\
                   \includegraphics[width=\linewidth]{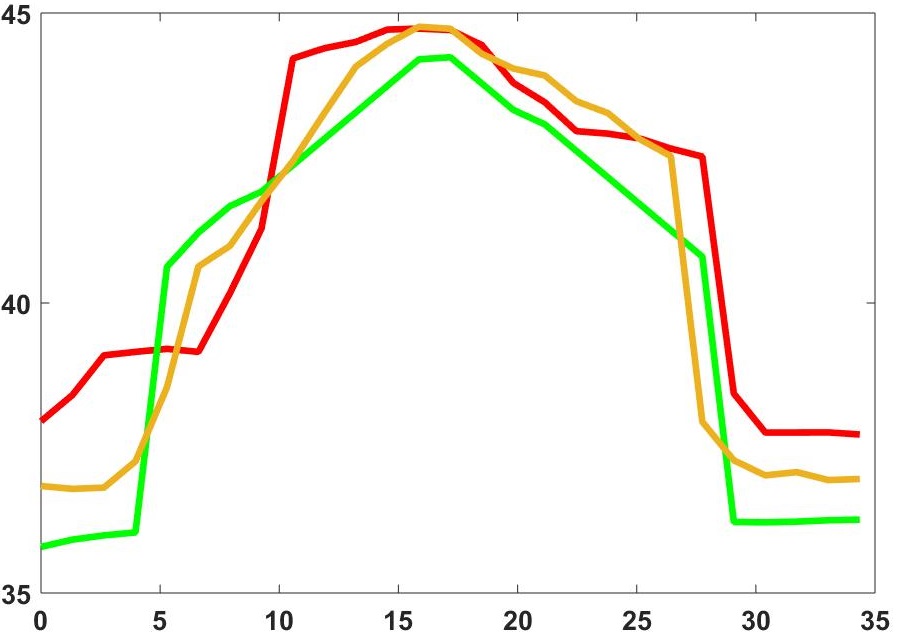}%
                 }%
                }
  \subcaptionbox{\WNetcGAN~\cite{Bittner19:new}\label{fig:IGARSS_profiles}}
                {\parbox{.3\linewidth}{%
                   \includegraphics[width=\linewidth]{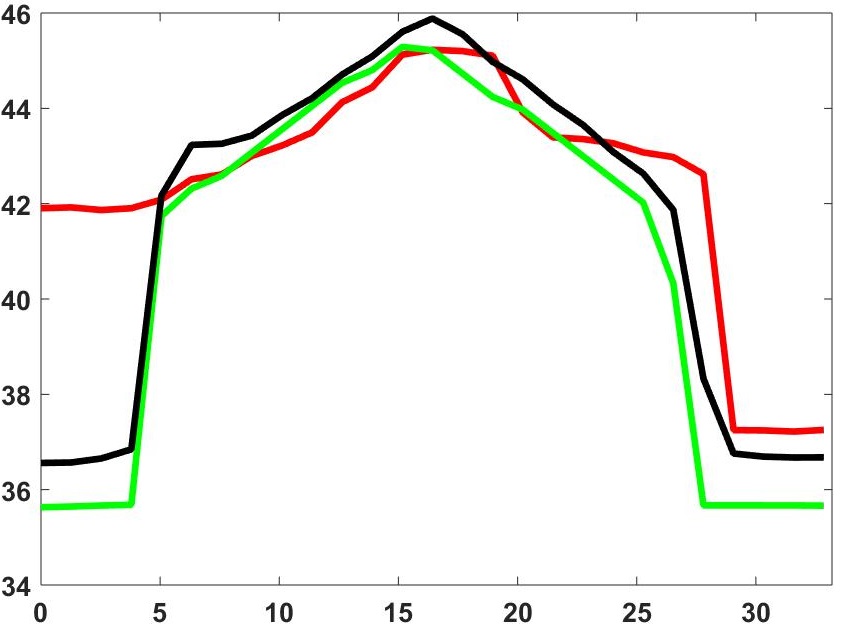}
                   \includegraphics[width=\linewidth]{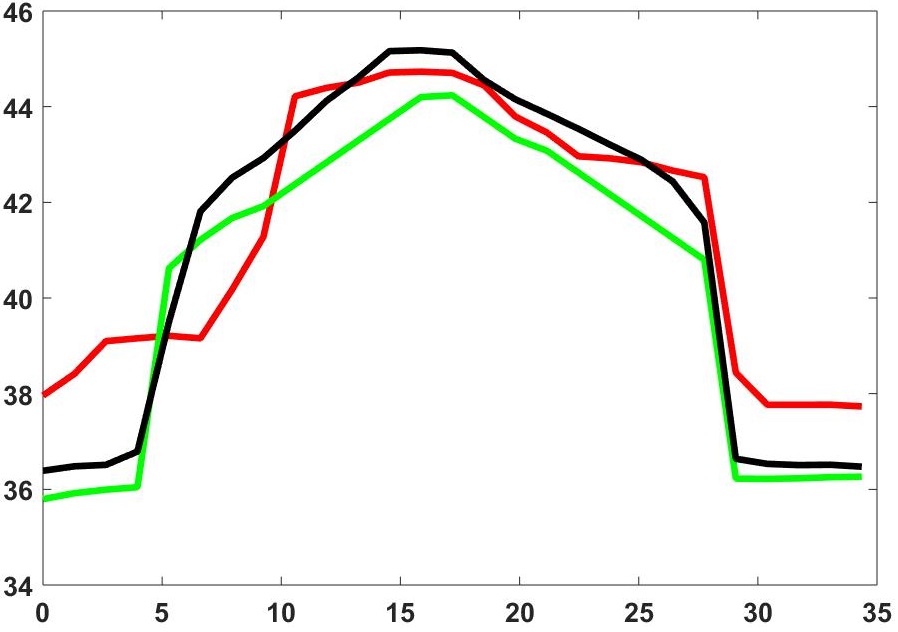}
                 }%
                }
  \subcaptionbox{\HybridcGAN (ours)\label{fig:HybridcGAN_profiles}}
                {\parbox{.3\linewidth}{%
                   \includegraphics[width=\linewidth]{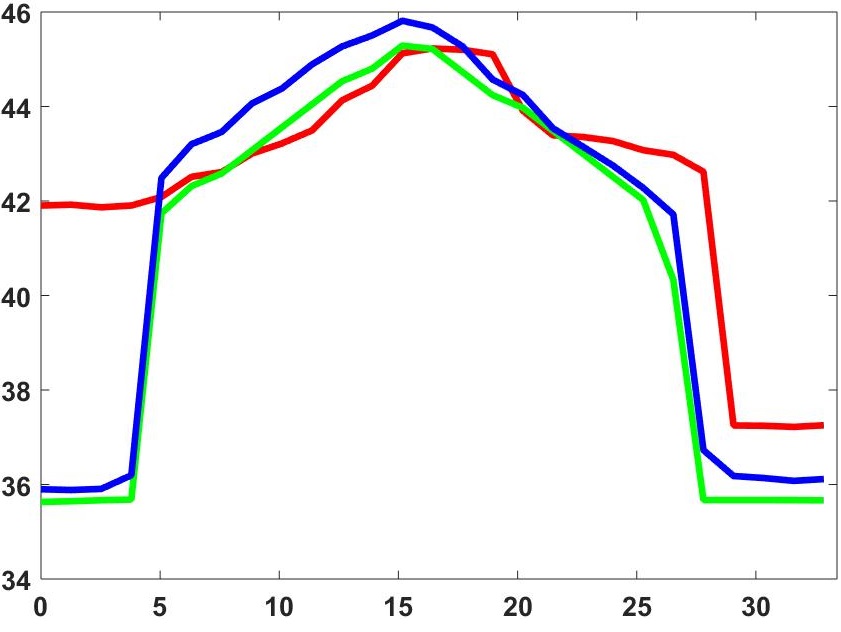}
                   \includegraphics[width=\linewidth]{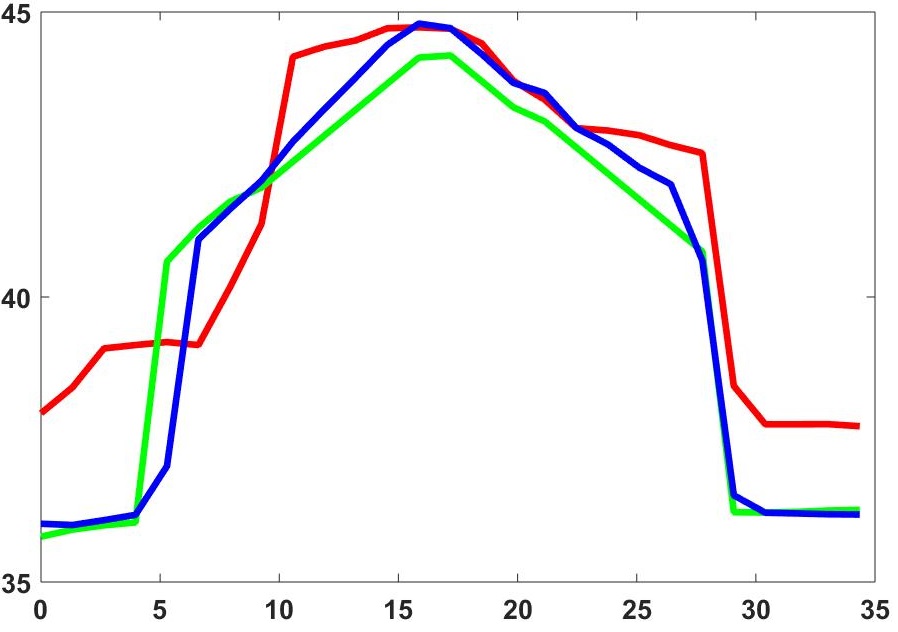}
                 }%
                }\\
  \vspace{0.2cm}
  %\fboxsep=0.5mm \fboxrule=0.3mm
  %\fcolorbox{black}{red}{\textcolor{red}{inp}}
  %\fcolorbox{white}{white}{input \gls{gl:DSM}} \quad
  %\fcolorbox{black}{yellow}{\textcolor{yellow}{inp}}
  %\fcolorbox{white}{white}{Single depth-stream \gls{gl:DSM}~\cite{bittner2018dsm}} \quad
  %\fcolorbox{black}{black}{\textcolor{black}{inp}}
  %\fcolorbox{white}{white}{\WNetcGAN~\cite{Bittner19:new}} \\
  %\fcolorbox{black}{blue}{\textcolor{blue}{inp}}
  %\fcolorbox{white}{white}{\HybridcGAN (ours)} \quad  
  %\fcolorbox{black}{green}{\textcolor{green}{inp}}
  %\fcolorbox{white}{white}{Ground Truth} \quad 
  \begin{center}
  \tikz{\path[draw=black,fill=red] (0,0) rectangle (0.5cm,0.4cm);}\hspace{0.1cm}\text{input \gls{gl:DSM}}\quad
  \tikz{\path[draw=black,fill=green] (0,0) rectangle (0.5cm,0.4cm);}\hspace{0.1cm}\text{Ground truth}\quad
  \tikz{\path[draw=black,fill=yellow] (0,0) rectangle (0.5cm,0.4cm);}\hspace{0.1cm}\text{Single depth-stream \gls{gl:DSM}~\cite{bittner2018dsm}}\\

  \tikz{\path[draw=black,fill=black] (0,0) rectangle (0.5cm,0.4cm);}\hspace{0.1cm}\text{\WNetcGAN~\cite{Bittner19:new}}\quad
  \tikz{\path[draw=black,fill=blue] (0,0) rectangle (0.5cm,0.4cm);}\hspace{0.1cm}\text{\HybridcGAN (ours)}
  \end{center}
  
  \caption{%
    Visual investigation of profiles for two selected buildings from the \glspl{gl:DSM} generated by 
    \subref{fig:cLSGAN_profile} the single-stream \gls{gl:cGAN} model~\cite{bittner2018dsm}, 
    \subref{fig:IGARSS_profiles} the two-stream \WNetcGAN model~\cite{Bittner19:new}, and
    \subref{fig:HybridcGAN_profiles} the proposed \HybridcGAN architecture. 
    The first row shows profiles of building \enquote{1} and the second row depicts the profiles of building \enquote{2} (\cf \cref{fig:test_O3andDSM2LODGEOKOLref_lsgan_NotSharedWeights_width_320_height_320}).
    %\newline{\huge\bfseries\color{red}Describe the meaning of the line colors!}
  }
  \label{fig:profiles}
\end{figure*}
\subsection{Implementation and Training Details}

The proposed \HybridcGAN network was realized using the \emph{PyTorch} python package based on the implementation published by \citet{isola2016image}.
The prepared training dataset covers \SI{353}{\square\kilo\meter} and consists of \num{21480} pairs of patches of size \SI{256x256}{\px} obtained by tiling the given satellite image on the fly with random overlap in both horizontal and vertical directions.
This procedure provides the network the possibility to learn building shapes, which, during one epoch, may be located on the patch border and, as a result, are only partially introduced to the network.
For the validation phase, an area covering \SI{6}{\square\kilo\meter} was used for tuning the hyper-parameters.
The \HybridcGAN network and others, used in this paper for comparison, were trained on minibatch \gls{gl:SGD} using the Adam optimizer~\cite{kingma2014adam} with an initial learning rate of $\alpha = 0.0002$ and momentum parameters $\beta_1 = 0.5$ and $\beta_2 = 0.999$. 
We set the weighting hyper-parameters $\lambda = 1000$ and $\gamma = 10$ after performing training and examining the resulting generated images from the validation dataset.
The complete number of epochs was set to 200 with a batch size of 5 on a single NVIDIA TITAN X (PASCAL) GPU with 12 GB of memory.

Different to the training phase, where two networks \generator and \discriminator were trained at the same time by alternating one gradient descent step of \discriminator and one gradient descent step of \generator, in inference phase only the trained generator model \generator was involved.
Stitching the predicted \gls{gl:LoD}2-like height images---entirely unseen during training and validation phases---with a fixed overlap of half the patch size in both horizontal and vertical directions, we generated the final full image covering an area of \SI{50}{\square\kilo\meter}.

\begin{figure*}[t]
  \centering
  \subcaptionbox{\label{fig:testArea_O_1}\Acrlong{gl:PAN} image}
                {\frame{\includegraphics[width=.32\linewidth]{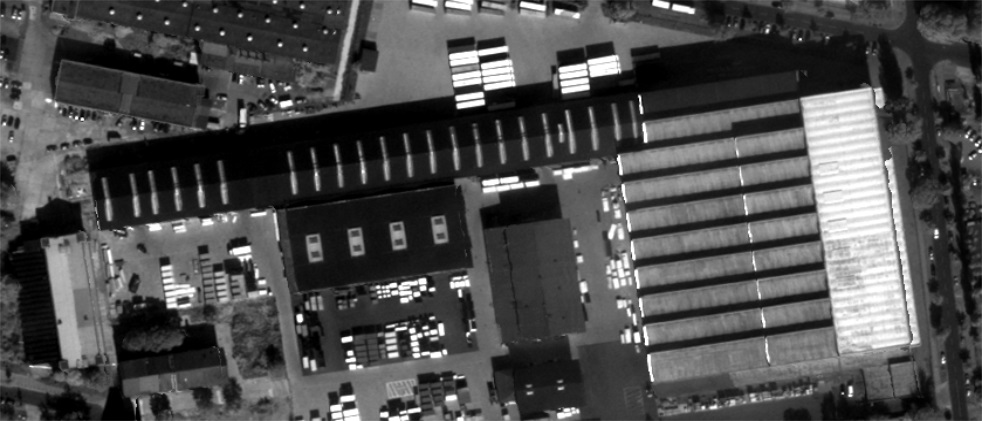}}
                }
  \subcaptionbox{\label{fig:testArea_Dfilled_geoid_1} Photogrammetric \gls{gl:DSM}}
                {\frame{\includegraphics[width=.32\linewidth]{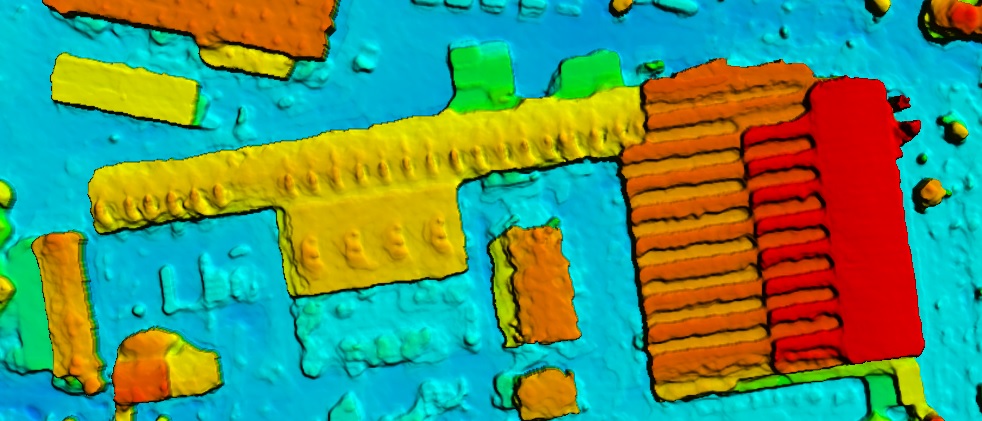}}
                }  
  \subcaptionbox{\label{fig:GT_1} Ground truth}
                {\frame{\includegraphics[width=.32\linewidth]{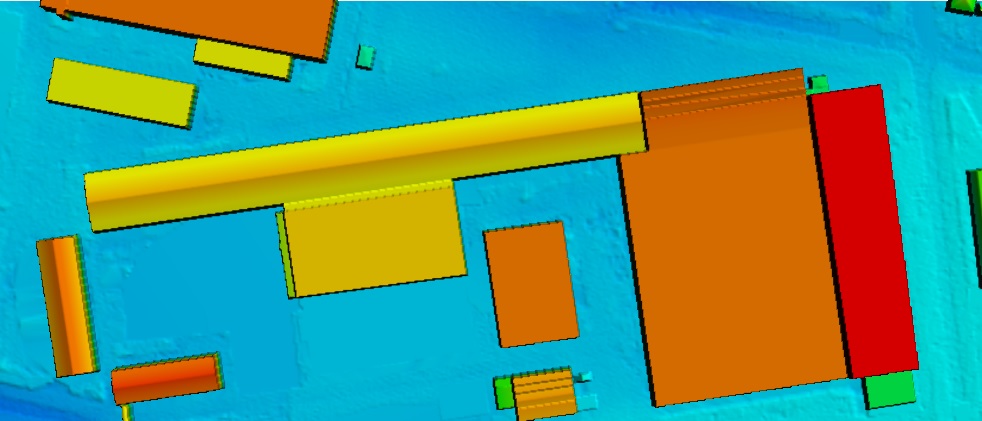}}
                }

  \subcaptionbox{\label{fig:test_DSM2LOD2GEOKOLref_wthTANH_1} Single depth-stream~\cite{bittner2018dsm} \gls{gl:DSM}}
                {\frame{\includegraphics[width=.32\linewidth]{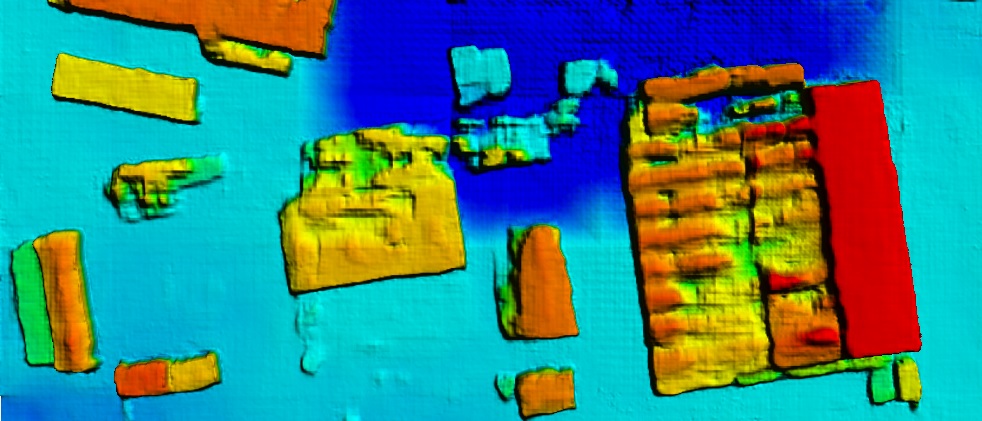}}
                }
  \subcaptionbox{\label{fig:test_OandDSM2LODGEOKOLref_1} \WNetcGAN~\cite{Bittner19:new}}
                {\frame{\includegraphics[width=.32\linewidth]{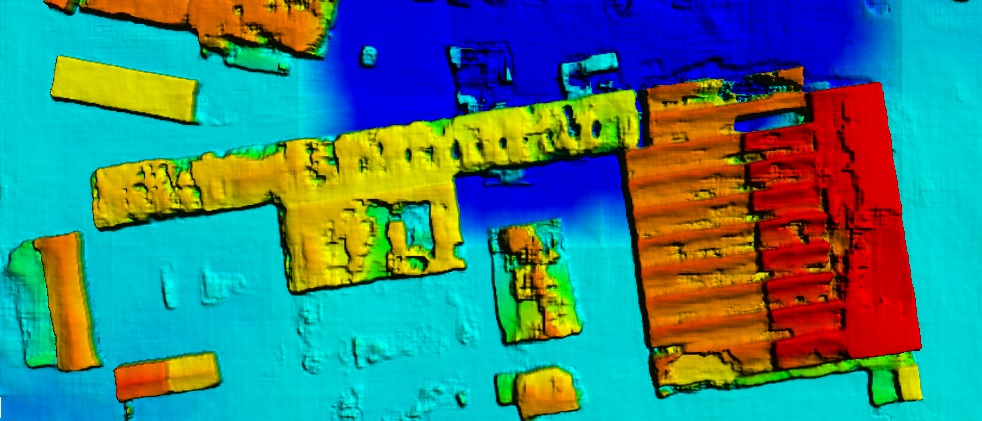}}
                } 
  \subcaptionbox{\label{fig:test_DSM2LOD2GEOKOLref_1} \HybridcGAN (ours)}
                {\frame{\includegraphics[width=.32\linewidth]{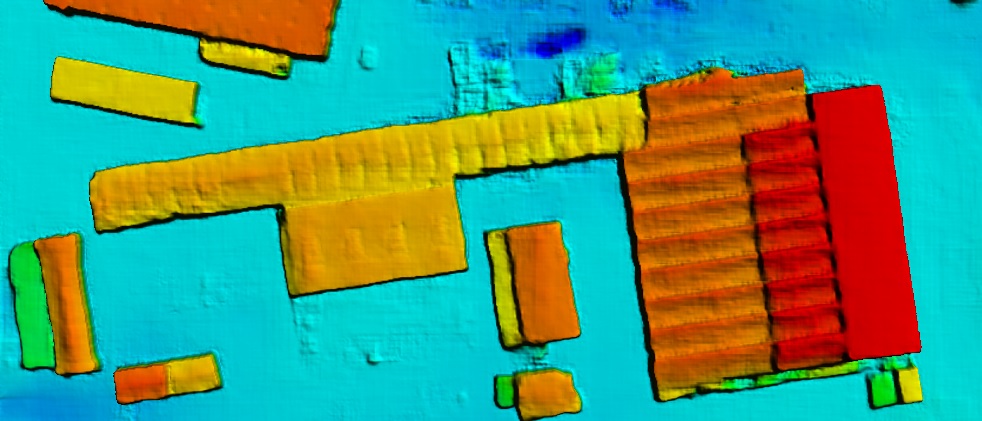}}
                }
  \caption{Visual analysis of \glspl{gl:DSM}, generated by 
  \subref{fig:test_DSM2LOD2GEOKOLref_wthTANH_1} the single-stream \gls{gl:cGAN} model~\cite{bittner2018dsm},  
  \subref{fig:test_OandDSM2LODGEOKOLref_1} the two-stream \WNetcGAN\cite{Bittner19:new}, and 
  \subref{fig:test_DSM2LOD2GEOKOLref_1} our proposed \HybridcGAN architecture. 
  The \gls{gl:DSM} images are color-shaded for better visualization. \label{fig:Area1}}
\end{figure*}

\begin{figure*}[t]
  \setlength{\fboxsep}{0pt}
  \centering
  \subcaptionbox{\label{fig:testArea_Ortho_2}\Gls{gl:PAN} image}
                {\fbox{\includegraphics[width=.32\linewidth]{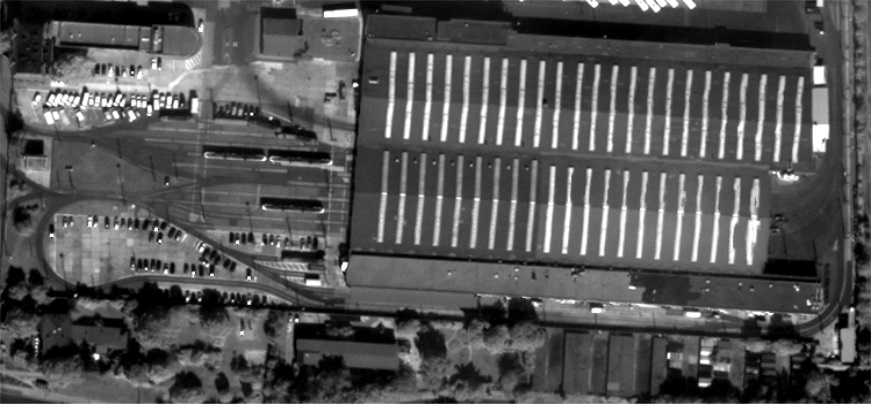}}
                }
  \subcaptionbox{\label{fig:testArea_Dfilled_geoid_2} Photogrammetric \gls{gl:DSM}}
                {\fbox{\includegraphics[width=.32\linewidth]{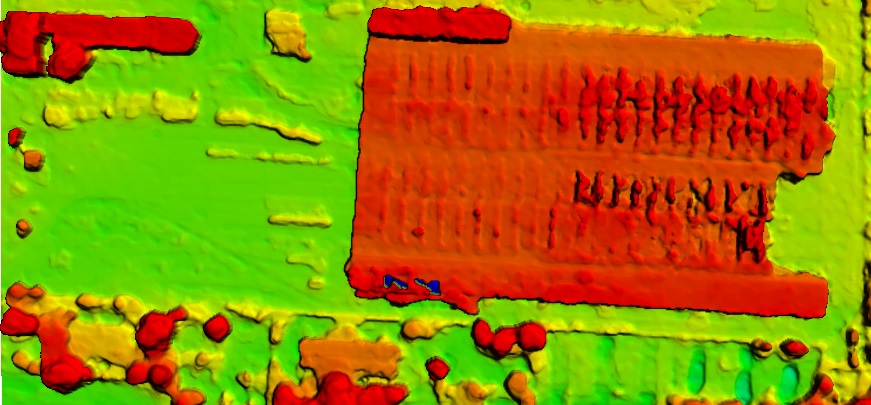}}
                }  
  \subcaptionbox{\label{fig:GT_2} Ground truth}
                {\fbox{\includegraphics[width=.32\linewidth]{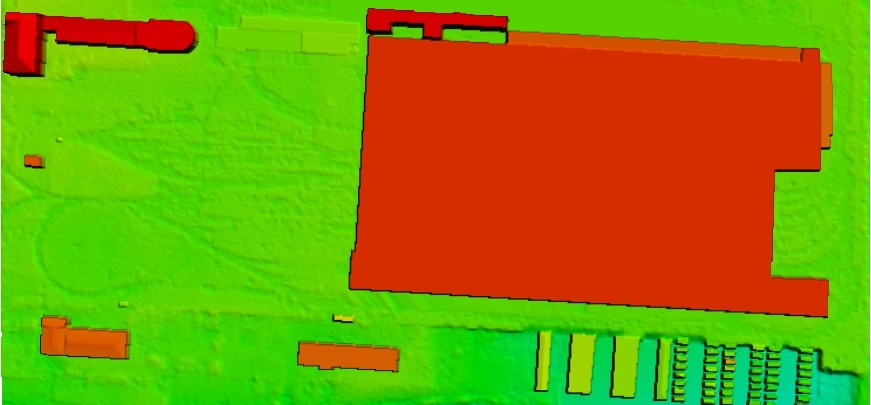}}
                }
                
  \subcaptionbox{\label{fig:DSM2LOD2GEOKOLref} Single depth-stream~\cite{bittner2018dsm}}
                {\fbox{\includegraphics[width=.32\linewidth]{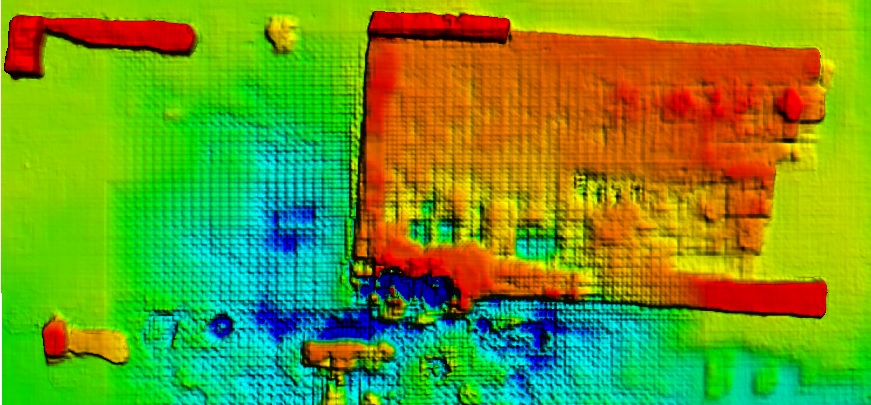}}
                }
  \subcaptionbox{\label{fig:test_O3andDSM2LODGEOKOLref_wnet_2} \WNetcGAN~\cite{Bittner19:new}}
                {\fbox{\includegraphics[width=.32\linewidth]{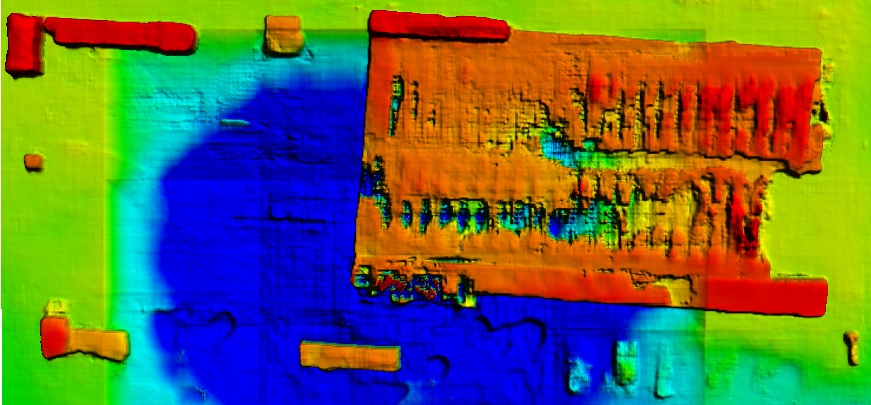}}
                } 
  \subcaptionbox{\label{fig:test_O3andDSM2LODGEOKOLref_lsgan_NotSharedWeights_2} \HybridcGAN (ours)}
                {\fbox{\includegraphics[width=.32\linewidth]{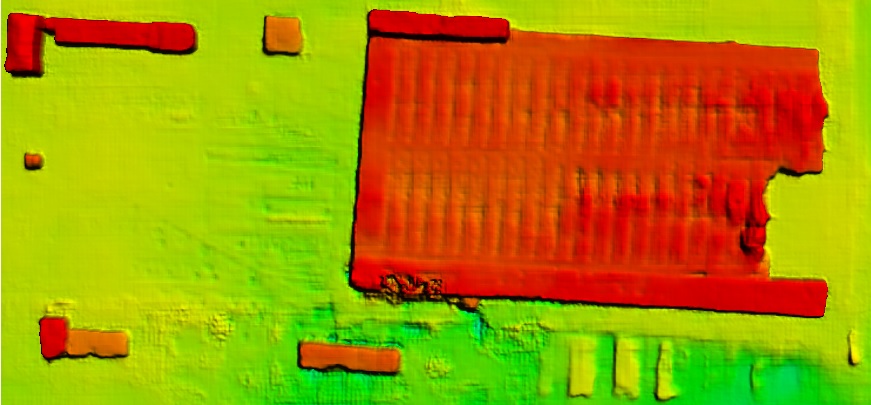}}
                }  
  \caption{%
  	Visual analysis of an \gls{gl:DSM}, generated by 
  	\subref{fig:DSM2LOD2GEOKOLref} the single-stream \gls{gl:cGAN} model~\cite{bittner2018dsm},  
  	\subref{fig:test_O3andDSM2LODGEOKOLref_wnet_2} the two-stream \WNetcGAN model~\cite{Bittner19:new}, and 
  	\subref{fig:test_O3andDSM2LODGEOKOLref_lsgan_NotSharedWeights_2} our proposed \HybridcGAN architecture. 
  	The \gls{gl:DSM} images are color-coded for better visualization.
  }
   \label{fig:Area2}
\end{figure*}

\section{Results and Discussion}

Selected test samples of the \glspl{gl:DSM} generated by the single-stream \gls{gl:cGAN} model~\cite{bittner2018dsm}, the two-stream \WNetcGAN model~\cite{Bittner19:new}, and our proposed \HybridcGAN model are illustrated in \cref{fig:smallBuildings,fig:Area1,fig:Area2}.
%\Cref{fig:smallBuildings,fig:Area1,fig:Area2} illustrate the selected test samples unseen during training. 
%Our previous works demonstrate that the \gls{gl:cGAN} models adapted for elevation model refinement are suit the task. 
One can notice that small buildings in all generated \glspl{gl:DSM} show more rectilinear borders and are not merged with adjacent trees, as present in the input \glspl{gl:DSM} (\cf \cref{fig:testArea_Dfilled_geoid_width_320_height_320,fig:testArea_Dfilled_geoid_1,fig:testArea_Dfilled_geoid_2}).
The integration of spectral information into the model obviously benefits the building reconstruction process. 
First of all, the number of reconstructed buildings is increased.  
For instance, the magenta arrows in \cref{fig:test_DSM2LOD2GEOKOLref_width_320_height_320} highlight the areas in the \gls{gl:DSM} generated by the single-stream \gls{gl:cGAN} model, where the model was not able to reconstruct individual buildings, as opposed to the \WNetcGAN network and our proposed \HybridcGAN network. 
Second, the roof ridge lines are more distinguishable and rectilinear.
This statement can be also confirmed by exemplary investigating the profiles of the two buildings highlighted in \cref{fig:test_O3andDSM2LODGEOKOLref_lsgan_NotSharedWeights_width_320_height_320}.
From \cref{fig:profiles} we can notice that the \HybridcGAN network was able to reconstruct much finer building shapes more similar to the ground truth. 
Moreover, the surfaces of roof planes are smoother or even flat in many cases affirming the influence of the normal vector loss.
The profiles also demonstrate the strength of all networks to separate the buildings from adjacent vegetation.
%The most of generated building shapes follow the correct pattern and have improved roof forms. 
%However, this statement is true for many residential and not big industrial buildings.
%How does the network behave in case of big and complicated building structures, which, it can happen, the network is not able to see as a whole within one patch? 
From the demonstrated results we can further conclude that most of the generated building shapes followed the correct pattern and feature improved roof forms. 
However, this statement is true for many residential and not big industrial buildings.
How does the network behave in case of big and complicated building structures? 
The single depth-stream \gls{gl:cGAN} model~\cite{bittner2018dsm} and the two-stream \WNetcGAN model~\cite{Bittner19:new} only partially extract such buildings. 
In case of spectral information integration, although at the late fusion, it helps to improve the silhouette of the buildings as well as the detailed constructions on the rooftops (\cf \cref{fig:test_O3andDSM2LODGEOKOLref_wnet_2,fig:test_OandDSM2LODGEOKOLref_1}) but still misses or has incompleted inside parts of structures.
%Moreover, as the input photogrammetric \glspl{gl:DSM} feature noise and many outlires, due to interpolation techniques, temporal changes or matching errors, those errors propagate along the height image reconstruction and influence the results (see dark blue areas in \cref{fig:test_DSM2LOD2GEOKOLref_wthTANH_1,fig:test_OandDSM2LODGEOKOLref_1}, and \cref{fig:DSM2LOD2GEOKOLref,fig:test_O3andDSM2LODGEOKOLref_wnet_2}).
Moreover, as the input photogrammetric \glspl{gl:DSM} contain noise and many outliers, they propagate along the height image reconstruction and influence the results, as indicated by the dark blue areas in \cref{fig:test_DSM2LOD2GEOKOLref_wthTANH_1,fig:test_OandDSM2LODGEOKOLref_1}, and \cref{fig:DSM2LOD2GEOKOLref,fig:test_O3andDSM2LODGEOKOLref_wnet_2}.
On the other hand, the proposed \HybridcGAN network was able to not only eliminate those artifacts, but also to reconstruct the complete building structures at any single detail.  
%From obtained results demonstrated in \cref{fig:test_DSM2LOD2GEOKOLref_1,fig:test_O3andDSM2LODGEOKOLref_lsgan_NotSharedWeights_2} one can also conclude that the building borders are more rectilinear that prove the positive influence of spectral information on building shape refinement. 
Besides, although the building rooftops seem to be entirely flat in the ground truth data (\cf \cref{fig:GT_1,fig:GT_2})---which is not the case in reality---, such cases do not confuse the model during the training phase and make it capable to preserve detailed 3D information from input photogrammetric \gls{gl:DSM}.
Those observations prove that the introduced \HybridcGAN architecture may successfully blend together the spectral and height information.
The earlier combination of both modalities forces the network to accommodate the information even better.

In order to evaluate the generated elevation models quantitatively, we utilized error metrics commonly used in the relevant literature~\cite{hohle2009accuracy,zhang2014geostatistical,elaksher2010refinement,hobi2012accuracy}, namely, %, such as \gls{gl:RMSE}, \gls{gl:NMAD}, \gls{gl:MAE}, 
the \gls{gl:RMSE}
\begin{align}
  \errorRMSE(\V{h},\hat{\V{h}}) &= \sqrt{\frac{1}{n}\sum\limits_{j=1}^n(\hat{h}_j - h_j)^2},
  \intertext{the \gls{gl:NMAD}}
  \errorNMAD(\V{h},\hat{\V{h}}) &= 1.4826 \cdot \underset{j}{\operatorname{median}} (|\Delta h_j - m_{\Delta \V{h}}|),
  \intertext{and the \gls{gl:MAE}}
  \errorMAE(\V{h},\hat{\V{h}}) &= \frac{1}{n}\sum\limits_{j=1}^n|\hat{h}_j - h_j|,
\end{align}
% 
% \begin{description}
% \item[RMSE:] $\varepsilon_\text{RMSE}(\V{h},\hat{\V{h}}) = \sqrt{\frac{1}{n}\sum\limits_{j=1}^n(\hat{h}_j - h_j)^2}$
% \item[NMAD:] $\varepsilon_\text{NMAD}(\V{h},\hat{\V{h}}) = 1.4826 \cdot \underset{j}{\operatorname{median}} (|\Delta h_j - m_{\Delta \V{h}}|)$ 
% \item[MAE:] $\varepsilon_\text{MAE}(\V{h},\hat{\V{h}}) = \frac{1}{n}\sum\limits_{j=1}^n|\hat{h}_j - h_j|$
% \end{description}
where 
$\V{h} = (h_j)_j$ and $\hat{\V{h}} = (\hat{h}_j)_j, 1 \leq j \leq n,$ denote the actually observed and the predicted heights, respectively,
height errors are defined as $\Delta h_j$, and the median error is $m_{\Delta \V{h}}$. 
The constant 1.4826, included in the \gls{gl:NMAD} metric, is comparable to the standard deviation if the data errors are distributed normally.
This estimator can be considered more robust to outliers in the dataset~\cite{hohle2009accuracy}.

To exclude the influence of time acquisition difference between the photogrammetric \glspl{gl:DSM} and the \gls{gl:CityGML} data model---carrying the risk of absence or the appearance of new buildings---, we %evaluated the generated \glspl{gl:DSM} on the presented three test areas and only for the buildings existed in both datasets.
%For this, we did not consider buildings, which exist either on photogrammetric \gls{gl:DSM} or on \gls{gl:LoD}2-\gls{gl:DSM}.
manually checked the evaluation regions in this regard and carried out evaluation on regions showing buildings in both the photogrammetric \gls{gl:DSM} as well as in the reference \gls{gl:LoD}2-\gls{gl:DSM}.

The obtained results for the three test area samples are presented in \cref{tab:table1,tab:table2,tab:table3}.
In comparison to the other methods, the \glspl{gl:DSM} created by the single-stream model~\cite{bittner2018dsm} showed inferior results in terms of $\varepsilon_\text{\acrshort{gl:RMSE}}$ for all three areas. 
As we have already pointed out before in \cref{fig:test_DSM2LOD2GEOKOLref_width_320_height_320}, the model was not able to reconstruct some buildings at all or only partially reconstructed them, exemplary highlighted in \cref{fig:test_DSM2LOD2GEOKOLref_wthTANH_1,fig:DSM2LOD2GEOKOLref}.

\begin{table}
  \caption{%
    Quantitative results for \gls{gl:RMSE}, \gls{gl:NMAD}, \gls{gl:MAE} metrics evaluated on 17 selected buildings existed on both input photogrammetric \gls{gl:DSM} and ground truth \gls{gl:LoD}2-\gls{gl:DSM} of the first area depicted in \cref{fig:smallBuildings}.%
  }
  \label{tab:table1}
  \centering
  \begin{tabular}{@{}v{3.75cm}xxx@{}} 
    \toprule
    \multicolumn{1}{@{}X}{Method} & \multicolumn{3}{X@{}}{Error} \\
    \cmidrule(l){2-4}
    & \multicolumn{1}{X}{\acrshort{gl:RMSE} (\si{\metre})}
    & \multicolumn{1}{X}{\acrshort{gl:NMAD} (\si{\metre})}
    & \multicolumn{1}{X@{}}{\acrshort{gl:MAE} (\si{\metre})}\\
    \cmidrule(r){1-1} \cmidrule(lr){2-2} \cmidrule(lr){3-3} \cmidrule(l){4-4} 
    \multicolumn{1}{@{}x}{photogrammetric \gls{gl:DSM}} 
    & 1.66 & 1.01 & 1.23 \\ 
    \multicolumn{1}{@{}x}{single-stream \gls{gl:cGAN} \cite{bittner2018dsm}} 
    & 2.28 & 1.09 & 1.86 \\ 
    \multicolumn{1}{@{}x}{\WNetcGAN\cite{Bittner19:new}} 
    & 1.63 & 0.72 & 1.22  \\
    \addlinespace
    \multicolumn{1}{@{}x}{\HybridcGAN (ours)} 
    & \textbf{1.52} & \textbf{0.62} & \textbf{0.96}   \\
    \bottomrule
  \end{tabular}
\end{table}

\begin{table}
  \caption{%
     Quantitative results for \gls{gl:RMSE}, \gls{gl:NMAD}, \gls{gl:MAE} metrics evaluated on 7 selected buildings existed on both input photogrammetric \gls{gl:DSM} and ground truth \gls{gl:LoD}2-\gls{gl:DSM} of the first area depicted in \cref{fig:Area1}.
  }
  \label{tab:table2}
  \centering
  \begin{tabular}{@{}v{3.75cm}xxx@{}} 
    \toprule
    \multicolumn{1}{@{}X}{Method} & \multicolumn{3}{X@{}}{Error} \\
    \cmidrule(l){2-4}
    & \multicolumn{1}{X}{\acrshort{gl:RMSE} (\si{\metre})}
    & \multicolumn{1}{X}{\acrshort{gl:NMAD} (\si{\metre})}
    & \multicolumn{1}{X@{}}{\acrshort{gl:MAE} (\si{\metre})}\\
    \cmidrule(r){1-1} \cmidrule(lr){2-2} \cmidrule(lr){3-3} \cmidrule(l){4-4} 
    \multicolumn{1}{@{}x}{photogrammetric \gls{gl:DSM}} 
    & 2.72 & \textbf{1.09} & \textbf{1.57}\\ 
    \multicolumn{1}{@{}x}{single-stream \gls{gl:cGAN} \cite{bittner2018dsm}} 
    & 4.13 & 1.88 & 2.68 \\ 
    \multicolumn{1}{@{}x}{\WNetcGAN\cite{Bittner19:new}} 
    & 3.89 & 2.03 & 2.64  \\
    \addlinespace
    \multicolumn{1}{@{}x}{\HybridcGAN (ours)} 
    & \textbf{2.64} & 1.34 & 1.69 \\
    \bottomrule
  \end{tabular}
\end{table}

\begin{table}
  \caption{%
     Quantitative results for \gls{gl:RMSE}, \gls{gl:NMAD}, \gls{gl:MAE} metrics evaluated on 4 selected buildings existed on both input photogrammetric \gls{gl:DSM} and ground truth \gls{gl:LoD}2-\gls{gl:DSM} of the first area depicted in \cref{fig:Area2}.
  }
  \label{tab:table3}
  \centering
  \begin{tabular}{@{}v{3.75cm}xxx@{}} 
    \toprule
    \multicolumn{1}{@{}X}{Method} & \multicolumn{3}{X@{}}{Error} \\
    \cmidrule(l){2-4}
    & \multicolumn{1}{X}{\acrshort{gl:RMSE} (\si{\metre})}
    & \multicolumn{1}{X}{\acrshort{gl:NMAD} (\si{\metre})}
    & \multicolumn{1}{X@{}}{\acrshort{gl:MAE} (\si{\metre})}\\
    \cmidrule(r){1-1} \cmidrule(lr){2-2} \cmidrule(lr){3-3} \cmidrule(l){4-4}
    \multicolumn{1}{@{}x}{photogrammetric \gls{gl:DSM}} 
    & 2.19 & 0.62 & \textbf{1.12}\\ 
    \multicolumn{1}{@{}x}{single-stream \gls{gl:cGAN} \cite{bittner2018dsm}} 
    & 3.38 & \textbf{0.45} & 2.99 \\ 
    \multicolumn{1}{@{}x}{\WNetcGAN\cite{Bittner19:new}} 
    & 3.16 & 1.15 & 2.02  \\
    \addlinespace
    \multicolumn{1}{@{}x}{\HybridcGAN (ours)} 
    & \textbf{1.91} & 0.56 &  1.22 \\
    \bottomrule
  \end{tabular}
\end{table}

With the intensity information integrated into the learning process, the \gls{gl:RMSE} error \errorRMSE decreased and, in case of our proposed \HybridcGAN, achieved the lowest value, even smaller than for the input photogrammetric \gls{gl:DSM}.
This observation provides evidence that the proposed model improves the noisy and inaccurate photogrammetric \glspl{gl:DSM} to a high level of details.
Considering the other two metrics, \HybridcGAN also outperformed the competing models, except for the third test area. 
The single-stream \gls{gl:DSM}~\cite{bittner2018dsm} achieved the lowest \gls{gl:NMAD} error with $\errorNMAD=0.45$. 
This seems reasonable, as the buildings show flat roofs in the ground truth, which is not the case for neither the input photogrammetric \gls{gl:DSM}, nor for the \glspl{gl:DSM} generated by the models with spectral information integrated.
As the \gls{gl:NMAD} metric is sensible to outliers, it shows higher values for the results with more detailed roofs.  

Our results demonstrate that deep learning models visually produce fairly reasonable reconstructed \glspl{gl:DSM}.
However, the used \gls{gl:RMSE}, \gls{gl:NMAD}, \gls{gl:MAE} metrics do not give good enough insight into the depth estimation quality, as they mainly consider the overall accuracy by reporting global statistics of depth residuals.
Moreover, the available ground truth data with only sufficient quality also influences evaluation procedure. 

\glsresetall
\section{Conclusion}

We presented a methodology for automatic building shape refinement from low-quality \glspl{gl:DSM} to the \gls{gl:LoD} 2 from multiple spaceborne remote sensing data on the basis of \glspl{gl:cGAN}. 
The network automatically combines the 
advantages of \gls{gl:PAN} imagery and photogrammetric \gls{gl:DSM}, while complementing their individual drawbacks, 
and, from obtained results, shows the potential of generating \glspl{gl:DSM} with completed not only residential but also industrial building structures.
Moreover, the generated roof surfaces are smoother and more planar, giving evidence of the positive influence of the auxiliary \emph{normal vector loss} function. 
Besides, a 3D visualization of the generated elevation models illustrates the realistic appearance of the buildings and their strong resemblance to the ground truth. 
%{\small
%\bibliographystyle{ieee}
%\bibliography{egbib}
%}

\printbibliography

\end{document}